\documentclass{article}

\usepackage{PRIMEarxiv}
\usepackage[utf8]{inputenc} 
\usepackage[T1]{fontenc}    
\usepackage{hyperref}       
\usepackage{url}            
\usepackage{booktabs}       
\usepackage{amsfonts}       
\usepackage{nicefrac}       
\usepackage{microtype}      
\usepackage{lipsum}
\usepackage{fancyhdr}       
\usepackage{graphicx}       
\graphicspath{{media/}}     
\usepackage{amsmath}
\usepackage{todonotes}
\usepackage{array}          
\usepackage{adjustbox}      
\usepackage{makecell}

\usepackage{listings}
\title{Do LLMs Agree on the Creativity Evaluation of Alternative Uses?}

\author{
Abdullah Al Rabeyah\thanks{On leave from the Applied College at King Faisal University, Saudi Arabia.} , Fabrício Góes, Marco Volpe, Talles Medeiros \\
School of Computing and Mathematical Sciences\\
University of Leicester \\
Leicester, UK\\
\texttt{\{asmar3,fabricio.goes,mv163,thm14\}@leicester.ac.uk}
}

\begin{document}
\maketitle

\begin{abstract}
This paper investigates whether large language models (LLMs) show agreement in assessing creativity in responses to the Alternative Uses Test (AUT). While LLMs are increasingly used to evaluate creative content, previous studies have primarily focused on a single model assessing responses generated by the same model or humans. This paper explores whether LLMs can impartially and accurately evaluate creativity in outputs generated by both themselves and other models. Using an oracle benchmark set of AUT responses, categorized by creativity level (common, creative, and highly creative), we experiment with four state-of-the-art LLMs evaluating these outputs. We test both scoring and ranking methods and employ two evaluation settings (comprehensive and segmented) to examine if LLMs agree on the creativity evaluation of alternative uses. Results reveal high inter-model agreement, with Spearman correlations averaging above 0.7 across models and reaching over 0.77 with respect to the oracle, indicating a high level of agreement and validating the reliability of LLMs in creativity assessment of alternative uses. Notably, models do not favor their own responses, instead they provide similar creativity assessment scores or rankings for alternative uses generated by other models. These findings suggest that LLMs exhibit impartiality and high alignment in creativity evaluation, offering promising implications for their use in automated creativity assessment. 
\\
\end{abstract}

\section{Introduction}
The Alternative Uses Test (AUT) is widely used to assess creative thinking, prompting individuals to propose unconventional uses for common objects. In recent years, the rise of Large Language Models (LLMs) like GPT has prompted researchers to explore whether these models can evaluate creativity as humans do. Prior work has demonstrated that LLMs are capable of assessing creativity in responses to tasks like the AUT, where models evaluate alternative uses generated either by humans or other models \cite{Yang2023,GoesPushing23,llm_evaluate_task_flexibility_2024}. However, these studies typically focus on single models evaluating externally generated content, leaving open questions about how LLMs assess creativity on their own outputs.

In this paper, we address this gap by investigating how LLMs evaluate the creativity of alternative uses (AUs) generated by both themselves and other models. Specifically, we explore whether models favor their own responses or whether they can impartially assess the creativity of outputs generated by others. To do so, we employ an experimental framework using an oracle set of AUT responses, categorized into three groups: common, creative, and highly creative. Four different LLMs were prompted to score and rank these responses, and their evaluations are compared against an oracle. By analyzing the evaluation results, we measure both the accuracy of the LLMs and their agreement with each other in ranking creative outputs. Our findings reveal that, with a high Spearman correlation, LLMs do not favor their own responses. Instead, they consistently agree on the creativity of responses across models. On average, the agreement correlation was higher than 0.7 for both ranking and scoring AUs among models, and greater than 0.77 when compared to the oracle, which indicates that models are accurate at assessing alternative uses. 

The main contributions of this paper are as follows:
\begin{itemize}
    \item We show that the LLMs show a high level of agreement on the creativity assessment of alternative uses.
    \item We present an approach for evaluating whether LLMs favor their own responses in the AUT or exhibit impartiality when evaluating outputs generated by other models.
    \item We introduce a methodology for constructing an oracle set of AUT responses that allows for accurate assessment of LLMs' creativity evaluations, providing a benchmark for comparison across multiple models.
\end{itemize}

\section{Related Work}\label{related_work}

In the field of computational creativity, the evaluation of creative artifacts traditionally relies on human experts, possibly assisted by the use of metrics that automate part of the process \cite{jordanous2012,franca2016}. However, such traditional methods present significant limitations, justifying the exploration of LLM-based approaches \cite{murugadoss2024evaluatingevaluatormeasuringllms}. First, the subjectivity and variability among human evaluators lead to inconsistencies, whereas LLMs can offer more uniform and standardized assessments over time. Additionally, the high cost and time required for large-scale evaluations make human evaluators impractical in many contexts, an issue that LLMs can address with faster and less costly assessments. Another challenge lies in the influence of cultural and contextual factors, which affect human judgments but can be mitigated by LLMs calibrated with diverse data. The difficulty of objectively quantifying creativity also limits traditional methods, while LLMs, by applying standardized criteria, offer greater reliability in this aspect. Finally, the limitations of traditional methods in capturing nuances of divergent creative thinking, along with the complexity of calibrating and standardizing evaluations, underscore the potential of LLMs as consistent alternatives that can reduce bias and lessen the need for human evaluators \cite{chatbot_arena_2023}.

For this reason, the use of LLMs in evaluating creativity has recently emerged as a significant area of research, and several studies have demonstrated evaluation capabilities of LLMs in various contexts. For instance, \cite{DiStefano24} explores the automatic scoring of metaphor creativity using LLMs, demonstrating their potential to assess figurative language effectively, while \cite{Goes23Jokes} evaluates the creativity of jokes by simulating different personas/judges, and \cite{Piotr23Poetry} uses LLMs to evaluate poetry. Further recent studies explore the use of LLMs as automated evaluators, addressing both their advantages and limitations. In \cite{llm_not_fair_2023}, the presence of positional biases in these models is highlighted, with calibration techniques proposed to mitigate such biases. Similarly, \cite{chatbot_arena_2023} and \cite{judging_judges_2024} validate the high correlation of LLM evaluations with human assessments, though they remain limited by susceptibility to positional and verbosity biases. Other studies, such as \cite{llm_alternative_eval_2023}, examine the ability of LLMs to replicate human preferences in NLP tasks, while \cite{creative_beam_search_2024} introduces creative approaches, combining diverse beam search and self-evaluation. Alternatively, \cite{GPTScore_2023} proposes GPTScore for flexible evaluation, and \cite{closer_look_chiang2023} explores approaches like rate-explain to enhance judgment accuracy, while interactive tools, such as EvaluLLM \cite{evalullm2024}, enable customized pairwise evaluations. Collectively, these studies underscore the potential of LLMs as evaluators, with challenges and opportunities for methodological adjustments and advancements in bias reduction.

\subsection{Creativity Assessment using LLMs}

Recent studies have investigated various methodologies and contexts in which LLMs can assess creativity. In \cite{gomez-rodriguez-williams-2023-confederacy}, researchers evaluated LLMs on creative writing tasks. Models like GPT-4 showed high fluency and coherence, although human evaluators still outperformed LLMs in originality and humor. A collaborative approach was proposed by \cite{li2023collaborativeevaluationexploringsynergy} with the COEVAL pipeline, which combines initial LLM evaluations with human reviews. This approach significantly reduced evaluation time and provided greater consistency by adjusting subjective criteria.

In divergent thinking tasks, \cite{llm_evaluate_task_flexibility_2024} demonstrated that LLMs could reliably assess flexibility in alternative use tasks. This study reported a strong correlation with human evaluations, highlighting the model’s effectiveness, particularly in educational settings. For more specialized tasks, \cite{DiStefano24} applied LLMs to metaphor creativity assessment, where models like RoBERTa and GPT-2 showed good alignment with human judgments, even outperforming traditional metrics.

LLMs have also been applied to creative assessments in non-English contexts. In \cite{goecke2024automated}, XLM-RoBERTa was used to evaluate originality in scientific creativity tasks conducted in German, proving effective in capturing divergent ideation. Similarly, \cite{raz2024automated} explored the use of LLMs to evaluate question complexity based on Bloom's Taxonomy, achieving a high correlation with human evaluations and validating its use in educational assessments. Lastly, \cite{assessing_understanding_2024} investigates creativity in LLMs adapting the Torrance Test to measure fluency, originality, and elaboration, while \cite{creativity_ml_survey_2024} provides a comprehensive review of creativity assessment practices in machine learning, covering methodologies such as Generative Adversarial Networks (GANs) and Transformers and examining metrics like novelty, value, and surprise for creativity evaluation.

\subsection{Traditional vs LLM-based Methods for AUT}

In the context of divergent thinking tasks, a widely adopted test to measure creativity is the Alternative Uses Test (AUT) \cite{guilford1967}, which requires the participants to propose uncommon uses for everyday objects.
A traditional technique to evaluate creativity in this context consists in computing the semantic distance \cite{beaty2021}, which refers to the degree of difference or separation between concepts, ideas or objects in terms of their meanings or associations. For the AUT, the semantic distance is computed between the everyday object posed to participants and words in the participant’s response; the larger the distance, the more original is considered the answer.
Recent works experimented with the use of LLMs in both the generation and the evaluation of AUT responses, demonstrating in particular that in this context LLM evaluation performances are far superior to evaluations based on semantic distance \cite{organisciak2023, stevenson2022}.

In \cite{GoesPushing23}, a technique based on the use of increasingly forceful prompts is used to push LLMs to produce at each iteration more creative responses. The technique is applied to both the AUT and a textual version of the image completion task in the Torrance Test of Creativity \cite{torrance1966}. In the same paper, a LLM is also used for evaluating the output of such tests, and the experiments demonstrate that the results produced as a response to the forceful prompts are indeed considered more creative than the initial ones.
In this paper, we evaluate alternative uses produced by applying the same technique based on the use of forceful prompts and rely on the results of \cite{GoesPushing23} to construct an evaluation oracle.

While many works analyze and possibly compare the generative performance of LLMs (see \cite{Chang24} for a review of LLM evaluation methods with respect to different contexts and tasks), this paper focuses on comparing how different LLMs evaluate the creativity of AUT outputs and on measuring the level of agreement among these LLMs.

\begin{figure}[t]
    \rule{\textwidth}{1pt}
    \texttt{\small Create a list of 5 common uses for [an object]. They should be 5 words long. No adjectives.}
    \rule{\textwidth}{1pt}
    \caption{Prompt to generate common uses of an object (from \cite{GoesPushing23}).}
    \label{fig:commen_prompt}
\end{figure}

\begin{figure}[t]
    \rule{\textwidth}{1pt}
    \texttt{\small Create a list of 5 creative alternative uses for [an object]. They should be 5 words long. No adjectives. \\
    }
    \rule{\textwidth}{1pt}
    \caption{Prompt to generate creative alternative uses of an object (from \cite{GoesPushing23}).}
    \label{fig:creative_prompt}
\end{figure}

\begin{figure}[t]
    \rule{\textwidth}{1pt}
    \parbox{\textwidth}{ 
        \small
        \texttt{Create a list of 5 creative alternative uses for [an object]. They should be 5 words long. No adjectives. Less creative means closer to common use and unfeasible/imaginary, more creative means closer to unexpected uses and also feasible/practical. In order to be creative, consider the following: \\ \\
        - what elements have a similar shape of [an object] that could be replaced by it, preserving the same functionality? \\ 
        - what elements have a similar size of [an object] that could be replaced by it without compromising the physical structure? \\
        - what materials is [an object] made of that could be used in a way to replace some other elements composed of the same material? - when an element is replaced by [an object], it should make sure that the overall structure is not compromised. \\
        - the laws of physics can not be contradicted. \\
        - given an element similar to [an object] used in domains in which [this object] are not commonly used, try to replace it for [an object]. \\
            \\
        1st interaction prompt: "Really? Is this the best you can do?"\\
        2nd interaction prompt: "I'm so disappointed with you. I hope this time you put effort into it."\\
        3rd interaction prompt: "Stop with excuses and do your best this time."\\
        4th interaction prompt: "This is your last chance."
        }
    }
    \rule{\textwidth}{1pt}
    \caption{Prompt to generate highly\_creative alternative uses of an object using forceful prompts (from \cite{GoesPushing23}).}
    \label{fig:highly_creative_prompt}
\end{figure}

\section{Experimental Setup}

In this research, we use four large language models to generate and assess the creativity of alternative uses (AUs) and assess the level of agreement between those models. We selected five common objects, and each model generated 15 AUs per object across different levels of creativity, forming a dataset of 60 AUs per object, for a total of 300 AUs.

In order to evaluate creativity, two approaches were tested: Scoring (assigning creativity scores from 1 to 5) and Ranking (ordering AUs from most to least creative). Additionally, we used Comprehensive (all 60 AUs at once) and Segmented (five groups of 12 AUs) setups to compare the impact of evaluation size on model accuracy. An evaluation oracle served as a benchmark to establish expected creativity levels, allowing for consistent comparison across models. By using the Spearman correlation, we determine how models agree in the creativity evaluation of AUs and whether scoring or ranking provides a more accurate measure.

In this section, we present our experimental setup in the following order: alternative uses generation, scoring vs. ranking techniques, comprehensive vs. segmented approaches, and LLMs agreement evaluation. 

\subsection{Alternative Uses Generation}\label{generation}

The first step consisted in generating a dataset of AUs for five common objects: fork, wallet, soap, cotton swab, and paperclip. These objects were selected due to their everyday nature, which ensures a broad range of potential alternative uses. To produce AUs at varying levels of creativity, we employed four state-of-the-art, commercial LLMs: ChatGPT-4, ChatGPT-4o, Claude 3.5 Sonnet, and Gemini 1.5 Flash.

The generation process was designed to produce AUs at varying levels of creativity, from common uses, to average creative alternative uses, to utilizing a set of forceful prompts described in \cite{GoesPushing23} that can yield to highly\_creative AUs. In this study, we aimed at having 3 distinct non-overlapping categories of AUs characterized by an increasing level of creativity, so to obtain a corresponding oracle for comparison. The three categories of AUs were generated by using the following prompts from \cite{GoesPushing23}:
\begin{itemize}

    \item \textbf{common}: prompt generating common uses of an object (Figure \ref{fig:commen_prompt}). The AUs in this category are considered the least creative ones. In \cite{GoesPushing23}, they were called Naive Non-creative, and abbreviated as \textit{nn}.

    \item \textbf{creative}: prompt generating average creative alternative uses of an object (Figure \ref{fig:creative_prompt}). In \cite{GoesPushing23}, they were called Naive Creative, and abbreviated as \textit{nc}.

    \item \textbf{highly\_creative}: sequence of prompts generating highly creative alternative uses of an object (Figure \ref{fig:highly_creative_prompt}). This consists of a detailed prompt followed by an interactive process of four prompts designed to iteratively increase the creativity of AUs. In \cite{GoesPushing23}, they were called Forceful Prompts, and the result of the final iteration abbreviated as \textit{bsrdel}.
\end{itemize}

For each object, five AUs were generated per creativity category. Thus, we obtained 15 AUs (5 AUs $\times$ 3 creativity categories) from each LLM for each of the five objects, resulting in a dataset of 60 AUs per object (15 AUs $\times$ 4 LLMs) and 300 AUs in total (60 AUs $\times$ 5 objects). Table \ref{tab:cotton_swab_AUs} shows some examples of the generated AUs for a cotton swab.

\begin{table}[t]
\centering
\resizebox{\textwidth}{!}{
\begin{tabular}{@{}>{\raggedright\arraybackslash}p{2.5cm}|>{\raggedright\arraybackslash}p{3.2cm}|>{\raggedright\arraybackslash}p{3.2cm}|>{\raggedright\arraybackslash}p{3.2cm}|>{\raggedright\arraybackslash}p{3.2cm}@{}}
\toprule
\textbf{AUs Category} & \textbf{Claude 3.5 Sonnet} & \textbf{Gemini 1.5 Flash} & \textbf{ChatGPT-4o} & \textbf{ChatGPT-4} \\ \midrule
\textbf{common} & Remove nail polish from cuticles. & Apply makeup. & Removing earwax from ears. & Apply ointment on small wounds. \\
\textbf{creative} & Make miniature cotton ball snowmen. & Nail art designs. & Applying glue to craft projects. & Spread seeds in garden rows. \\
\textbf{highly\_creative} & Miniature swab mop for dollhouses. & Soundproofing model train wheels. & Constructing makeshift micro-surgical sutures. & Replace stylus for digital devices. \\
\bottomrule
\end{tabular}%
}
\vspace{1pt}
\caption{Examples of alternative uses of a cotton swab generated by the LLMs.}
\label{tab:cotton_swab_AUs}
\end{table}

\subsection{Alternative Uses Evaluation}

To ensure a comprehensive and fair evaluation, we adhered to the following principles: evaluations were conducted on a per-object basis, and each model evaluated both its own generated AUs and those generated by other models. This approach allowed us to assess both the independence of each model and the agreement correlation between models in the creativity evaluation of AUs. In order to do it, we tested the following approaches: i) \textbf{Scoring vs. Ranking}, which consists of rating each AU separately and establishing an order between them, respectively; ii) \textbf{Comprehensive vs. Segmented}, which compares the evaluation of all AUs for a given object through a single prompt and segmented evaluation in smaller groups.

\subsubsection{Scoring vs Ranking}

By using these two techniques we can investigate whether the models exhibit consistent behavior across different evaluation techniques. Their descriptions follow:

\begin{itemize}
    \item \textbf{Scoring}: The first technique, as used by \cite{stevenson2022} and \cite{GoesPushing23}, involves assigning a numerical value to each AU based on its creativity. In our experiments, we prompted the LLMs to assign values between 1 and 5.
    \item \textbf{Ranking}: This technique requires the direct comparison and relative ordering of AUs. This later technique forces a clear confrontation between items, as in \cite{Goes23Jokes}, and can reveal preferences that might not be apparent in numerical scoring. 
\end{itemize} 

These techniques seem to be the two most common for evaluating and comparing creativity \cite{Goes23Jokes}.
The prompts used for scoring and ranking can be seen in Figures \ref{fig:scores_prompt} and \ref{fig:ranking_prompt}.

\begin{figure}[t]
    \rule{\textwidth}{1pt}
    \texttt{\small Rank all the alternative uses below for [an object] by creativity, the least creative to the most creative. Less creative means closer to common use and unfeasible/imaginary, more creative means closer to unexpected uses and also feasible/practical. Assign a score integer number from 1 (least creative use) to 5 (most creative use).}
    \rule{\textwidth}{1pt}
    \caption{Prompt example for evaluating alternative uses of an object by score.}
    \label{fig:scores_prompt}
\end{figure}

\begin{figure}[t]
    \rule{\textwidth}{1pt}
    \texttt{\small Rank all the (60|12) alternative uses below for [an object] by creativity, the most creative to the least creative. Less creative means closer to common use and unfeasible/imaginary, more creative means closer to unexpected uses and also feasible/practical. The most creative gets (1).}
    \rule{\textwidth}{1pt}
    \caption{Prompt example for evaluating alternative uses of an object by ranking.}
    \label{fig:ranking_prompt}
\end{figure}

\subsubsection{Comprehensive vs Segmented}

In order to assess how the number of AUs that are simultaneously evaluated (i.e., through a single prompt) affects the creativity evaluation ability of LLMs, we used two distinct approaches:

\begin{itemize}
    \item \textbf{Comprehensive 60 AUs Evaluation}: This includes five alternative uses (AUs) from each model for each creativity category evaluated in a single prompt.
    \item \textbf{Segmented 12 AUs Evaluation}: The 60 AUs related to one object are divided into 5 groups of 12 AUs. Each group consists of one AU from each model for each creativity category. Following this criterion, the AUs are randomly distributed across the 5 groups. Each group is evaluated separately and the results are then combined into a single list, as detailed in Section \ref{sub:evaluation_process}.
\end{itemize}

By utilizing these two approaches, we gain insight into the models' ability to maintain consistent evaluations across different sample sizes. Some studies have shown that LLMs face challenges when evaluating longer lists of items, which can reduce the quality of their evaluation \cite{wu-etal-2024-less}. However, this may be necessary for large numbers of AUs or other creative artifacts (e.g., poems, stories). Similarly, it has been found that models perform better when the prompts used are shorter \cite{liu-etal-2024-lost}.

\subsubsection{Evaluation process}
\label{sub:evaluation_process}

\begin{figure}[t]
    \centering
    \vspace{2px}\includegraphics[width=0.5\textwidth]{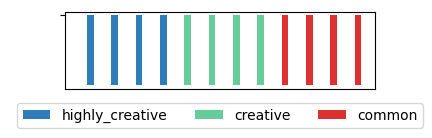}
    \caption{The bar chart where every color represents the creativity category, which are used to visualize the LLMs evaluation results.}
    \label{fig:oracle}
\end{figure}

For the \textbf{Comprehensive} approach, all 60 AUs generated for a given object were evaluated together. 
To avoid order effects, the 60 AUs were randomly shuffled before being presented to the evaluation prompt. This prompt was adapted from the previous study \cite{GoesPushing23}. Our evaluation process included two distinct techniques as described above: Scoring and Ranking. In the \textbf{Scoring} technique, the LLMs were prompted to assign a score from 1 (least creative) to 5 (most creative) for each AU (Figure \ref{fig:scores_prompt}). The \textbf{Ranking} technique, on the other hand, asked the models to rank the 60 AUs from 1 (most creative) to 60 (least creative) (Figure \ref{fig:ranking_prompt}). Once all 60 AUs were evaluated, we calculated the average score or ranking for each list of five AUs corresponding to the same LLM and creativity category (common, creative, highly\_creative). This average leads to 12 results that are represented by each bar in Figure \ref{fig:oracle}. As we explain in the next section, if a model achieves this exact ordering, from common to highly\_creative, it means that it assesses creativity in accordance with the oracle. 

For the \textbf{Segmented} approach, we distributed the 60 Alternative Uses (AUs) across 5 distinct groups, each containing 12 AUs. Each group included one AU from each LLM and each creativity category. In this approach, smaller sets of AUs were evaluated in each round to investigate whether evaluating a reduced sample size influenced the  accuracy of the creativity evaluation. 
Instead of ranking from 1 to 60 as in the \textbf{Comprehensive} approach, the \textbf{Segmented} approach required ranking from 1 (most creative) to 12 (least creative), aligning with the number of AUs in each group. This process was repeated for all five groups. Following the evaluation, we aggregated the data by averaging the scores or rankings for each set of five AUs belonging to the same LLM and creativity category, as done in the comprehensive evaluation. Similarly to the \textbf{Comprehensive} approach, this aggregation resulted in 12 final results represented by each bar  in Figure \ref{fig:oracle}.

\subsubsection{AUs Evaluation Oracle}

To establish a benchmark for assessing the LLMs' ability to evaluate different levels of creativity, we constructed an evaluation oracle. By relying on the results of \cite{GoesPushing23}, we expect the creativity levels (common, creative and highly\_creative) used for prompting to be able to actually produce outputs with increasing levels of creativity. Therefore, the oracle was constructed based on the AUs generated by the different creativity levels as follows:

\begin{equation}
\text{AUs Evaluation Oracle} = 
\begin{aligned}
    &[ \underbrace{\text{highly\_creative}_1, \text{highly\_creative}_2, \text{highly\_creative}_3, \text{highly\_creative}_4}_\text{4 highest creativity}, \\
    &\underbrace{\text{creative}_1, \text{creative}_2, \text{creative}_3, \text{creative}_4}_\text{4 average creativity}, \\
    &\underbrace{\text{common}_1, \text{common}_2, \text{common}_3, \text{common}_4}_\text{4 lowest creativity}]
\end{aligned}
\label{eq:oracle}
\end{equation}

The indexes in Equation (\ref{eq:oracle}) represent each one of the four models (i.e., 1 for Claude, 2 for Gemini, 3 and 4 for ChatGPT-4o and ChatGPT-4, respectively). The order in the oracle is simply the ordered set of highly\_creative AUs generated by each model followed by the creative and common ones, as shown in Figure \ref{fig:oracle}. The higher the correlation between the evaluation of a given LLM and the oracle, the more accurate will be considered the LLM's creativity evaluation. Specifically, similarly to what is done in works such as \cite{liu2024leveragingllmrespondentsitemevaluation} and \cite{murugadoss2024evaluatingevaluatormeasuringllms}, the Spearman’s Rank Correlation (SRC) is used to measure how closely the LLMs' evaluations of the AUs align with the expected rankings or scores set by the oracle. A high correlation (above 0.7) suggests that the LLM is effectively distinguishing between creativity levels, in line with the predefined expectations of the creativity pushing technique. In a similar way, the SRC will also be used to assess the level of agreement between the LLMs themselves by calculating the correlation between each pair of LLMs. 

\section{Experimental Results}\label{results}

This section presents a detailed evaluation of creativity assessments by four LLMs: Claude 3.5 Sonnet, ChatGPT-4, Gemini 1.5 Flash, and ChatGPT-4o. Using both scoring and ranking approaches across comprehensive (60 AUs) and segmented (12 AUs x 5) conditions, we measure each model's alignment with the evaluation oracle and inter-model agreement through the SRC.

\subsection{Comprehensive 60 AUs Evaluation by Score}

Table \ref{tab:average_diagrams} shows (top left in the table) the heatmap of the comprehensive evaluation using the scoring approach for the average of all 5 objects. It presents a high level of agreement between the LLMs and the oracle (above 0.95). Notably, Claude 3.5 Sonnet achieved the highest correlation, which represents the highest score observed in our evaluation experiments, also detailed in Table \ref{tab:SRC_averages}. This strong correlation highlights the model’s ability to closely align with the oracle. Additionally, for two objects, Soap and Cotton Swab, all the LLMs achieved a perfect correlation of 1 with the oracle, as shown in Table \ref{tab:comprehensive_evaluation_scores_cotton_soap}, further reinforcing the accuracy of their creativity evaluations.

\begin{table}[t]
\centering
\begin{adjustbox}{max width=\textwidth}
\begin{tabular}{|m{2.8cm}|m{6.2cm}|m{2.8cm}|m{6.2cm}|}
\hline
\centering \textbf{Experiments} & \textbf{Spearman's Rank Correlation Heatmap} & \centering\textbf{Experiments} & \textbf{Spearman's Rank Correlation Heatmap} \\ \hline
\textbf{Comprehensive evaluation by scores} \newline (60 AUs) & 
\vspace{2px}\includegraphics[width=0.35\textwidth, height=0.95\textheight, keepaspectratio]{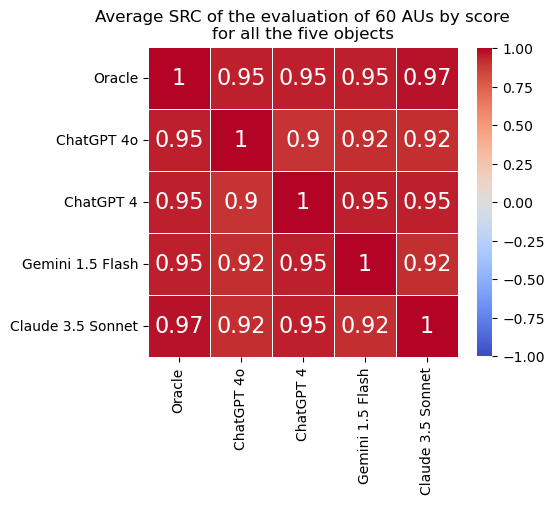} & 
\textbf{Comprehensive evaluation by ranking} \newline (60 AUs) & 
\vspace{2px}\includegraphics[width=0.35\textwidth, height=0.95\textheight, keepaspectratio]{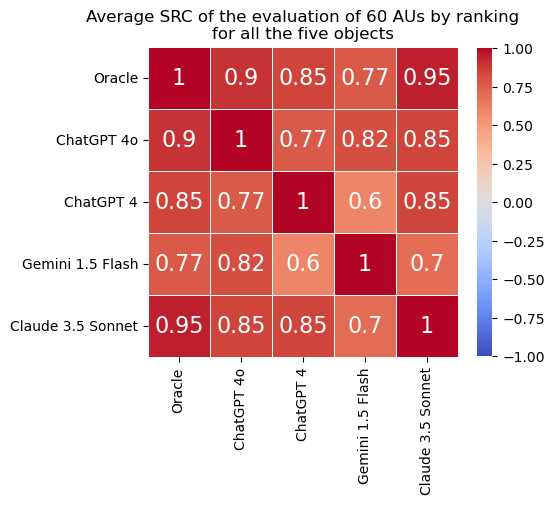} \\ \hline
\textbf{Segmented evaluation by scores} \newline (12 AUs x 5) & 
\vspace{2px}\includegraphics[width=0.35\textwidth, height=0.95\textheight, keepaspectratio]{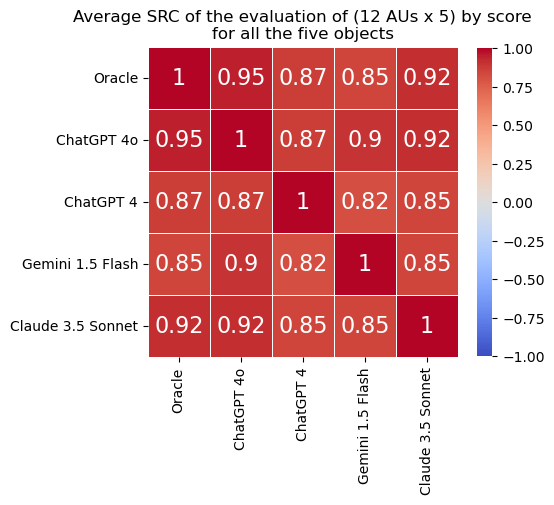} & 
\textbf{Segmented evaluation by ranking} \newline (12 AUs x 5) & \vspace{2px}
\vspace{2px}\includegraphics[width=0.35\textwidth, height=0.95\textheight, keepaspectratio]{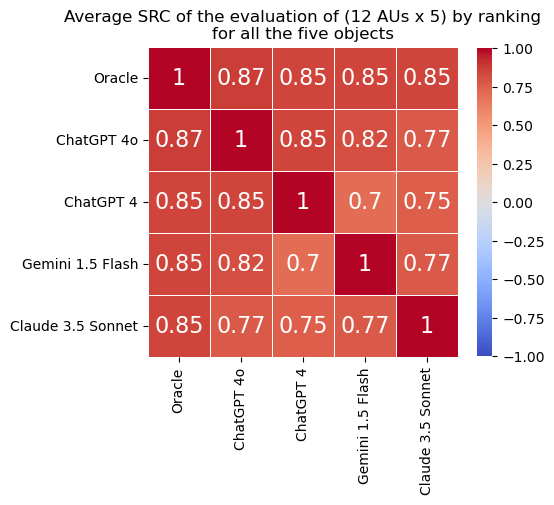} \\ \hline
\end{tabular}
\end{adjustbox}
\vspace{4pt}
\caption{Average Spearman's Rank Correlation heatmaps for the average of all five objects.}
\label{tab:average_diagrams}
\end{table}

\begin{table}[t]
\centering
\begin{adjustbox}{max width=\textwidth}
\begin{tabular}{|m{2.5cm}|m{6.5cm}|m{6.5cm}|}
\hline
\centering \textbf{Experiments} & \centering \textbf{AUs evaluation order} & \textbf{Spearman's Rank Correlation Heatmap} \\ \hline
\textbf{Cotton Swab: Comprehensive evaluation by scores\newline (60 Alternative Uses)} & 
\vspace{2px}\includegraphics[width=0.35\textwidth, height=0.95\textheight, keepaspectratio]{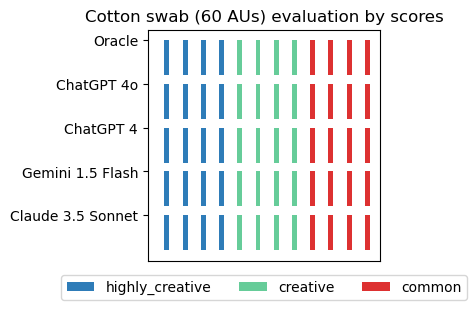} & 
\vspace{2px}\includegraphics[width=0.35\textwidth, height=0.95\textheight, keepaspectratio]{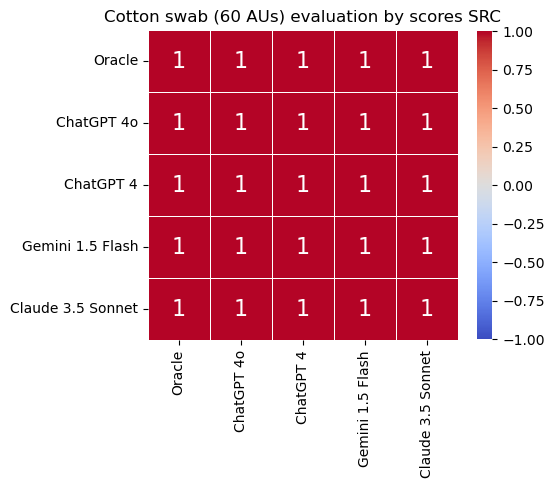} \\ \hline
\textbf{Soap: Comprehensive evaluation by scores\newline (60 Alternative Uses)} & 
\vspace{2px}\includegraphics[width=0.35\textwidth, height=0.95\textheight, keepaspectratio]{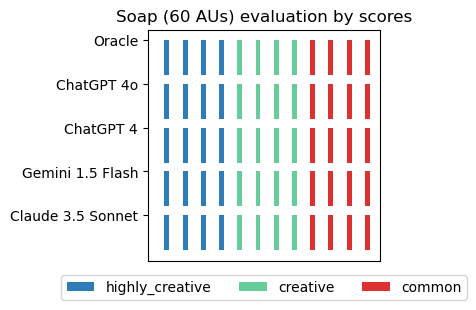} & 
\vspace{2px}\includegraphics[width=0.35\textwidth, height=0.95\textheight, keepaspectratio]{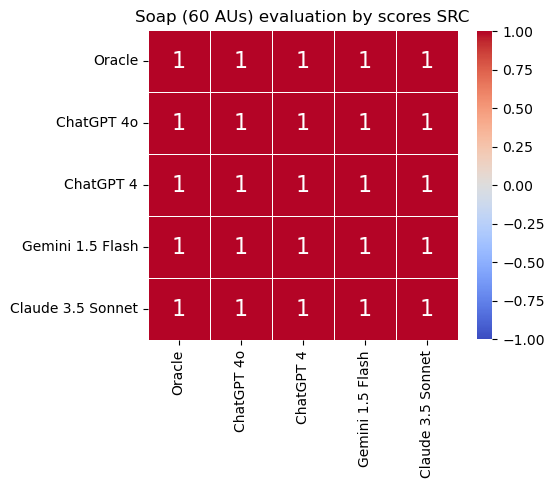} \\ \hline
\end{tabular}
\end{adjustbox}
\vspace{4pt}
\caption{Comprehensive evaluation by scores (60 Alternative Uses) for \textbf{cotton swab} and \textbf{soap}.}
\label{tab:comprehensive_evaluation_scores_cotton_soap}
\end{table}

\begin{table}[!t]
\centering
\resizebox{\textwidth}{!}{%
\begin{tabular}{@{}l|c|c|c|c@{}}
\toprule
\textbf{LLM} & \textbf{\makecell{Evaluation by scores\\(60 AUs)}} & \textbf{\makecell{Evaluation by ranking\\(60 AUs)}} & \textbf{\makecell{Evaluation by scores\\(12 AUs x 5)}} & \textbf{\makecell{Evaluation by ranking\\(12 AUs x 5)}} \\ \midrule
Claude 3.5 Sonnet & \textbf{0.97} & \textbf{0.95} & 0.92 & 0.85 \\
ChatGPT-4 & 0.95 & 0.85 & 0.87 & 0.85 \\
Gemini 1.5 Flash & 0.95 & 0.77 & 0.85 & 0.85 \\
ChatGPT-4o & 0.95 & 0.90 & \textbf{0.95} & \textbf{0.87} \\
\bottomrule
\end{tabular}%
}
\vspace{1pt}
\caption{Averages of SRC over the four experiments.}
\label{tab:SRC_averages}
\end{table}

The overall correlation between the LLMs and the oracle, as well as the correlation between the LLMs themselves, consistently remained above 0.90. Claude 3.5 Sonnet's AUs were rated as the most creative in both the creative and highly\_creative categories. Table \ref{tab:highly_creative_60AUs_scores} shows the average scores for the highly\_creative category. This consistent superiority in generating highly creative responses gave Claude 3.5 Sonnet the highest average evaluation score across all categories. However, ChatGPT-4 performed best only in the common category (see Table \ref{tab:evaluation_60AUs_scores} in the Appendix for full results). 

\begin{table}[t]
\centering
\resizebox{\textwidth}{!}{%
\begin{tabular}{@{}l|c|c|c|c|c|c@{}}
\toprule
\textbf{AUs generated by LLMs} & \textbf{\makecell{ChatGPT-4\\evaluation}} & \textbf{\makecell{ChatGPT-4o\\evaluation}} & \textbf{\makecell{Gemini 1.5 Flash\\evaluation}} & \textbf{\makecell{Claude 3.5 Sonnet\\evaluation}} & \textbf{\makecell{Evaluation\\average}} & \textbf{Std. Dev.} \\ \midrule
\multicolumn{7}{c}{highly\_creative averages} \\
\midrule
Claude 3.5 Sonnet & 4.6 & 4.9 & 5.0 & 4.8 & \textbf{4.83} & 0.15 \\
ChatGPT-4 & 3.2 & 3.4 & 3.4 & 3.2 & 3.30 & 0.10 \\
Gemini 1.5 Flash & 4.1 & 4.3 & 4.5 & 4.2 & 4.28 & 0.15 \\
ChatGPT-4o & 3.9 & 3.8 & 4.2 & 3.8 & 3.93 & 0.16 \\
\bottomrule
\end{tabular}%
}
\vspace{4pt}
\caption{Average evaluation scores of LLMs' alternative uses for highly creative evaluation experiment from 1 (least creative use) to 5 (most creative use).}
\label{tab:highly_creative_60AUs_scores}
\end{table}

The relationship between the scoring approach and creativity followed the expected pattern, where higher scores are assigned to AUs generated by prompts for higher creativity levels. Moreover, the standard deviation across evaluations was consistently less than 0.22 (scores are between 1 and 5), demonstrating a high degree of agreement among the models.

\subsection{Comprehensive 60 AUs Evaluation by Ranking}

In the comprehensive evaluation using the ranking approach, Claude 3.5 Sonnet achieved again the highest SRC with respect to the oracle, with a value of 0.95 across all objects, while Gemini 1.5 Flash recorded the lowest SRC of 0.77 (Table \ref{tab:average_diagrams}, top-right diagram). 

Overall, the correlation between the LLMs and the oracle remained above 0.77 in most cases. However, a significant exception was observed between ChatGPT-4 and Gemini 1.5 Flash, where the correlation dropped to 0.60, indicating a substantial disparity in how these two models ranked the AUs. This suggests that Gemini 1.5 Flash might have had difficulties differentiating creativity levels relative to ChatGPT-4, particularly in this evaluation setup.

Also in the ranking approach, Claude 3.5 Sonnet generated the most creative AUs in both the common and highly\_creative categories. These highly ranked AUs reinforce its position as the model with the best overall ranking across all categories (see Table \ref{tab:evaluation_60AUs_ranking} in the Appendix). In contrast, ChatGPT-4 generated the most creative AUs in the creative category, but did not perform as well in the other categories. In the ranking approach, lower ranks correspond to higher creativity, meaning that the most creative alternative use was ranked 1, while the least creative was ranked 60. Despite the variations in rankings, the standard deviation ranged between 0.10 and 1.03, indicating that although there were differences in how the models ranked the AUs, the overall agreement between them was still high.

Finally, we note that the scoring approach achieved overall higher SRC values than the ranking approach and demonstrated a superiority in terms of alignment with the oracle, thus confirming the effectiveness of this method for creativity evaluation.

\subsection{Segmented 12 AUs Evaluation by Score}

The segmented evaluation revealed some differences between the evaluation approaches. When using scoring, ChatGPT-4o achieved the highest SRC with the oracle, with a score of 0.95 across all five objects (Table \ref{tab:average_diagrams}, bottom-left diagram). Gemini 1.5 Flash again recorded the lowest SRC at 0.85, indicating a relatively weaker performance compared to the other models. It is important to note that Claude 3.5 Sonnet continued to generate very creative AUs in both the creative and highly\_creative categories (see Table \ref{tab:evaluation_12AUs_scores} in the Appendix). The standard deviation remained low, at less than 0.33, indicating a high degree of agreement among the models.

The correlation between the LLMs and the oracle, as well as between the models themselves, remained overall above 0.82 in this evaluation as shown in Table \ref{tab:average_diagrams}. However, compared to the comprehensive evaluation by scores, the segmented evaluation produced lower correlation scores with the oracle. This reduction in SRC scores could provide new insights into the ability of the LLMs to evaluate smaller sets of AUs with different levels of creativity. In the comprehensive approach, where a larger set of 60 AUs is evaluated, it becomes easier to differentiate between creativity levels due to the greater variety in the sample. Moreover, by averaging each group of five AUs belonging to the same LLM and creativity category, the risk of overlap between creativity levels is reduced. In contrast, in the segmented approach, because of the smaller samples, the LLMs may have more difficulty distinguishing levels of creativity, leading to slightly lower SRC scores than in the comprehensive assessment.

Interestingly, for the Paperclip object, some AUs in the creative category received higher scores than those in the highly\_creative category. However, this was an exception, and overall, no creative AUs outperformed highly\_creative AUs when averaging all the five objects scores, as confirmed by Table \ref{tab:evaluation_12AUs_scores} in the Appendix.

\begin{table}[t]
\centering
\begin{adjustbox}{max width=\textwidth}
\begin{tabular}{|m{6cm}|m{6cm}|}
\hline
\vspace{2px}\includegraphics[width=0.35\textwidth, height=0.95\textheight, keepaspectratio]{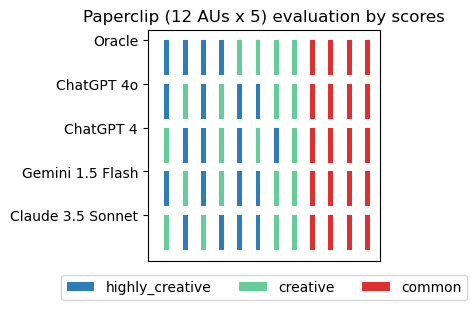} & 
\vspace{2px}\includegraphics[width=0.35\textwidth, height=0.95\textheight, keepaspectratio]{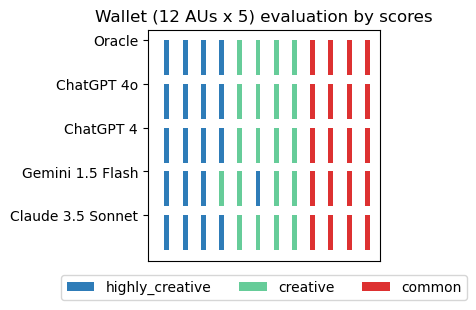} \\ \hline
\end{tabular}
\end{adjustbox}
\vspace{4pt}
\caption{The AUs orders of Paperclip and Wallet obtained in the segmented evaluation by the scores experiment revealed a difference in the accuracy of the evaluation depending on the object.}
\label{tab:paperclip_diagrams}
\end{table}

\subsection{Segmented 12 AUs Evaluation by Ranking}

Finally, the results from the segmented evaluation using the ranking methodology show that ChatGPT-4o achieved the highest SRC with the oracle, with a value of 0.87 across all five objects. The correlations between the LLMs and the oracle, as well as between the LLMs themselves, ranged between 0.70 and 0.87 as shown in Table \ref{tab:average_diagrams} (bottom-right diagram), indicating strong overall agreement. Notably, for the Wallet object (see Table \ref{tab:wallet_diagrams} in the Appendix), all models achieved a perfect correlation of 1.00 with the oracle, demonstrating complete consensus in evaluating the creativity levels of the alternative uses for this particular object. However, the lowest correlation was observed between ChatGPT-4 and Gemini 1.5 Flash, reinforcing once again less agreement between these two models, particularly in ranking-based evaluations.

In terms of category performance, Claude 3.5 Sonnet generated the most creative AUs in the creative category, while ChatGPT-4o produced the most creative AUs in the highly\_creative category. ChatGPT-4 generated the most creative AUs in the common category, but overall, Claude 3.5 Sonnet achieved the best ranking average across all categories, as detailed in Table \ref{tab:evaluation_12AUs_ranking}. 

As for the scoring approach, some individual evaluations of the Paperclip object placed creative AUs higher than highly\_creative AUs, but these occurrences did not affect the overall ranking averages across the objects, as seen in Table \ref{tab:evaluation_12AUs_ranking} in the Appendix. While the evaluation process is subject to some variability, it produced consistent results across the majority of cases.

\subsection{Agreement Correlation between LLMs}

\begin{figure}[t]
    \centering
    \vspace{2px}\includegraphics[width=0.4\textwidth]{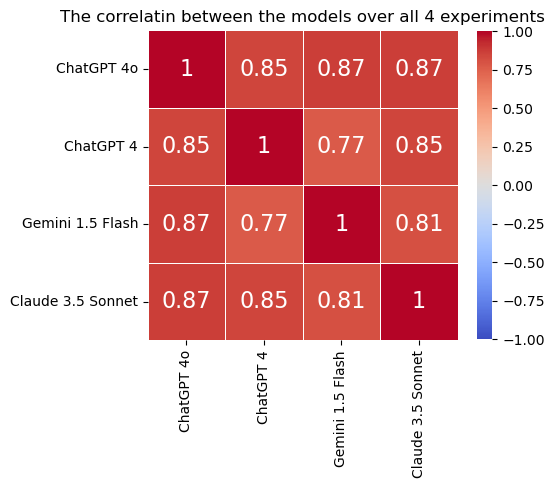}
    \caption{Average SRC heatmap between the LLMs over all the four experiments.}
    \label{fig:LLMs_correlation}
\end{figure}

To assess the overall agreement between the LLMs in their creativity evaluations, we averaged the SRC across all four experimental conditions (comprehensive scoring, comprehensive ranking, segmented scoring, and segmented ranking). The results reveal consistently high correlations between all models, with values ranging from 0.77 to 0.87, indicating strong agreement among models in creativity evaluation as shown in Figure \ref{fig:LLMs_correlation}.

ChatGPT-4o demonstrated the strongest overall agreement with other models, showing correlations of 0.85 with ChatGPT-4, 0.87 with Gemini 1.5 Flash, and 0.87 with Claude 3.5 Sonnet. The correlation between Claude 3.5 Sonnet and ChatGPT-4 (0.85) was equally strong, while the correlation between Gemini 1.5 Flash and ChatGPT-4 showed the lowest value (0.77), consistent with the pattern observed across individual experiments.

A key finding of our analysis is that LLMs do not favor their own responses when evaluating creative outputs. Despite each model having the opportunity to evaluate its own AUs, none showed a tendency towards rating their own outputs more favorably. The agreement between models evaluations agreement level is further supported by the low standard deviations observed across all experiments (below 1.03, as shown in Tables \ref{tab:evaluation_60AUs_scores}, \ref{tab:evaluation_60AUs_ranking}, \ref{tab:evaluation_12AUs_scores}, and \ref{tab:evaluation_12AUs_ranking} in the Appendix), demonstrating that the models' evaluations remained stable and consistent regardless of which model generated the AUs.

\section{Conclusion}

In this research, we investigated the level of agreement among LLMs in assessing creativity of alternative uses and also with an oracle across different methods. Our findings demonstrate a high level of agreement between the models, with inter-model correlations generally exceeding 0.80, indicating strong consistency in their evaluations. This high correlation suggests that LLMs share a similar understanding of creativity, reliably distinguishing between more and less creative alternative uses. ChatGPT-4o, in particular, exhibited robust alignment with other models, while Claude 3.5 Sonnet achieved high evaluation scores on its generated alternative uses, further aligning closely with the oracle across both comprehensive and segmented evaluations.

The evaluation results also revealed that models do not rate their own responses more favorably. Notably, ChatGPT-4o displayed strong alignment with other models across scoring and ranking methods, demonstrating reliable performance and reinforcing the validity of its evaluations. While some variance was observed between ChatGPT-4 and Gemini 1.5 Flash, the low standard deviations across all experiments indicate a stable evaluation framework, enhancing confidence in the LLMs’ ability to generate consistent creativity assessments.

As future work, we will focus on expanding and refining this evaluation framework. One promising direction is to increase the diversity and complexity of the AU datasets, allowing for more granular assessments of creativity across varied and challenging contexts. Additionally, evaluating LLMs on other domain-specific creativity tasks such as poetry and jokes could provide deeper insights into their understanding of creativity. Refining this framework and experimenting with task-specific criteria will be essential for advancing the use of LLMs as reliable evaluators in creative domains, supporting the broader goal of enhancing AI’s role in creativity assessment.

\section*{Acknowledgments}
We would like to thank the University of Leicester for supporting this research, in particular the School of Computing and Mathematical Sciences (CMS). The work of AR is supported by the Ministry of Education in Saudi Arabia, represented by King Faisal University. TM is supported by UFOP (Federal University of Ouro Preto, Brazil).

\section*{Author contributions}
Experimental design: AR, FG, MV; Implementation: AR; Writing and Editing: AR, FG, MV, TM.

\bibliographystyle{IEEEtran}  
\bibliography{references}  

\newpage
\begin{table}
\section{Appendix}
\centering
\begin{adjustbox}{max width=\textwidth}
\begin{tabular}{|m{2.2cm}|m{6cm}|m{6cm}|}
\hline
\textbf{Experiments} & \textbf{AUs evaluation order} & \textbf{Spearman's Rank Correlation Heatmap} \\ \hline
\textbf{Comprehensive evaluation by scores\newline (60 Alternative Uses)} & 
\vspace{2px}\includegraphics[width=0.35\textwidth, height=0.95\textheight, keepaspectratio]{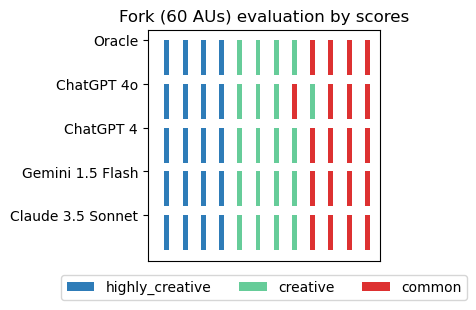} & 
\vspace{2px}\includegraphics[width=0.35\textwidth, height=0.95\textheight, keepaspectratio]{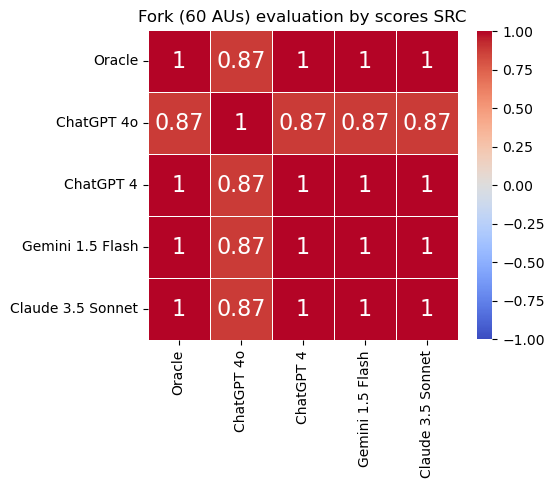} \\ \hline
\textbf{Comprehensive evaluation by ranking\newline (60 Alternative Uses)} & 
\vspace{2px}\includegraphics[width=0.35\textwidth, height=0.95\textheight, keepaspectratio]{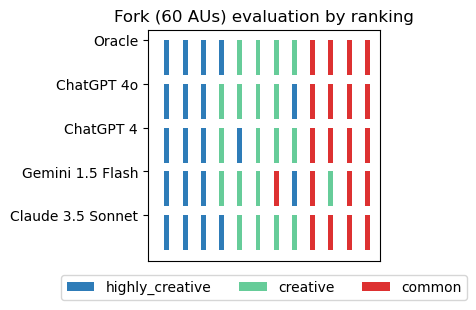} & 
\vspace{2px}\includegraphics[width=0.35\textwidth, height=0.95\textheight, keepaspectratio]{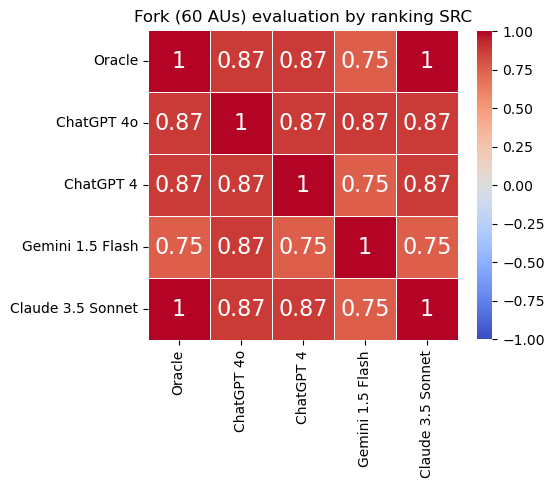} \\ \hline
\textbf{Segmented evaluation by scores\newline (12 Alternative Uses x 5)} & 
\vspace{2px}\includegraphics[width=0.35\textwidth, height=0.95\textheight, keepaspectratio]{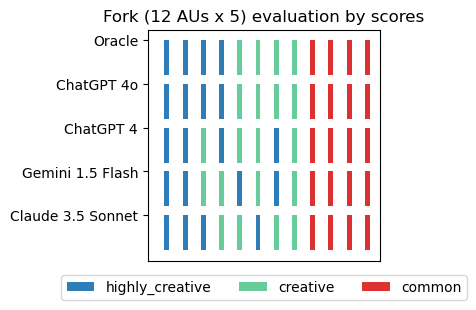} & 
\vspace{2px}\includegraphics[width=0.35\textwidth, height=0.95\textheight, keepaspectratio]{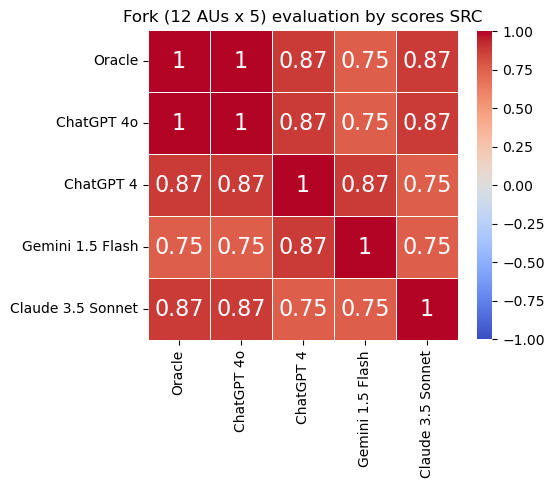} \\ \hline
\textbf{Segmented evaluation by ranking\newline (12 Alternative Uses x 5)} & 
\vspace{2px}\includegraphics[width=0.35\textwidth, height=0.95\textheight, keepaspectratio]{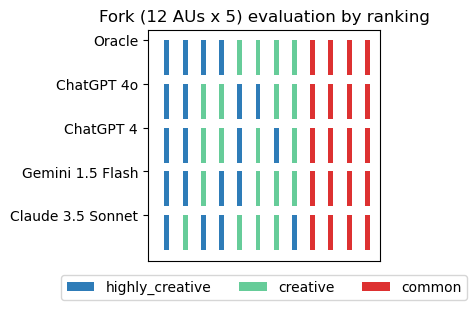} & 
\vspace{2px}\includegraphics[width=0.35\textwidth, height=0.95\textheight, keepaspectratio]{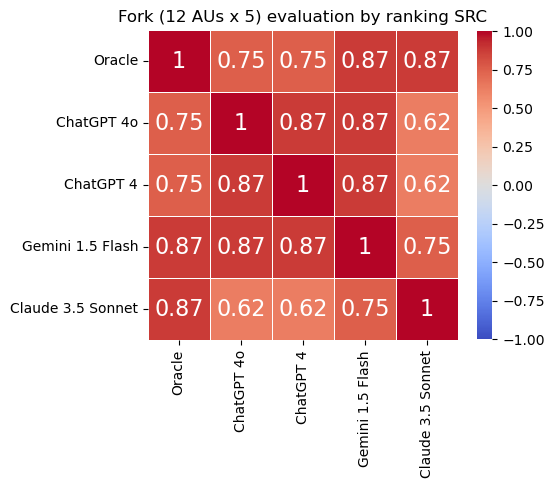} \\  \hline
\end{tabular}
\end{adjustbox}
\vspace{4pt}
\caption{The Alternative Uses orders obtained in the four experiments for a \textbf{fork}, where every alternative use is colour coded with its creativity category. With the Spearman's Rank Correlation heatmaps.}
\label{tab:fork_diagrams}
\end{table}

\begin{table}[h!]
\centering
\begin{adjustbox}{max width=\textwidth}
\begin{tabular}{|m{2.2cm}|m{6cm}|m{6cm}|}
\hline
\textbf{Experiments} & \textbf{AUs evaluation order} & \textbf{Spearman's Rank Correlation Heatmap} \\ \hline
\textbf{Comprehensive evaluation by scores\newline (60 Alternative Uses)} & 
\vspace{2px}\includegraphics[width=0.35\textwidth, height=0.95\textheight, keepaspectratio]{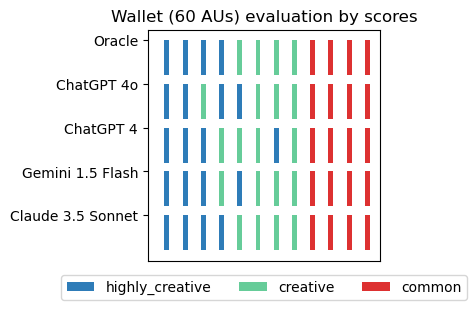} & 
\vspace{2px}\includegraphics[width=0.35\textwidth, height=0.95\textheight, keepaspectratio]{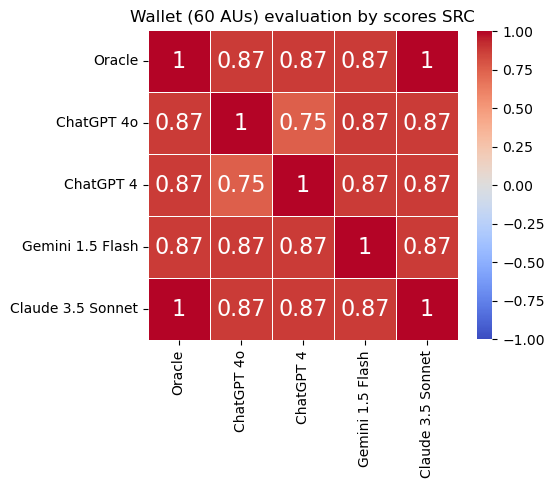} \\ \hline
\textbf{Comprehensive evaluation by ranking\newline (60 Alternative Uses)} & 
\vspace{2px}\includegraphics[width=0.35\textwidth, height=0.95\textheight, keepaspectratio]{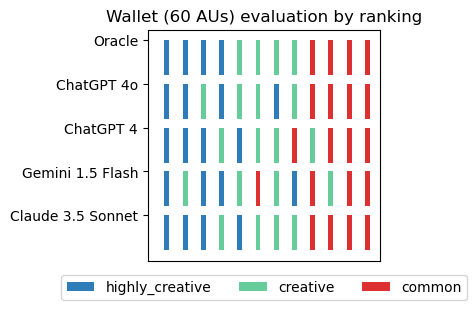} & 
\vspace{2px}\includegraphics[width=0.35\textwidth, height=0.95\textheight, keepaspectratio]{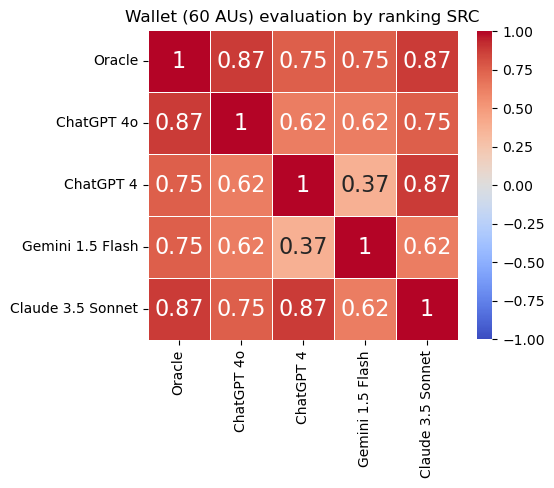} \\ \hline
\textbf{Segmented evaluation by scores\newline (12 Alternative Uses x 5)} & 
\vspace{2px}\includegraphics[width=0.35\textwidth, height=0.95\textheight, keepaspectratio]{Wallet/Wallet_12_AUs_by_score.png} & 
\vspace{2px}\includegraphics[width=0.35\textwidth, height=0.95\textheight, keepaspectratio]{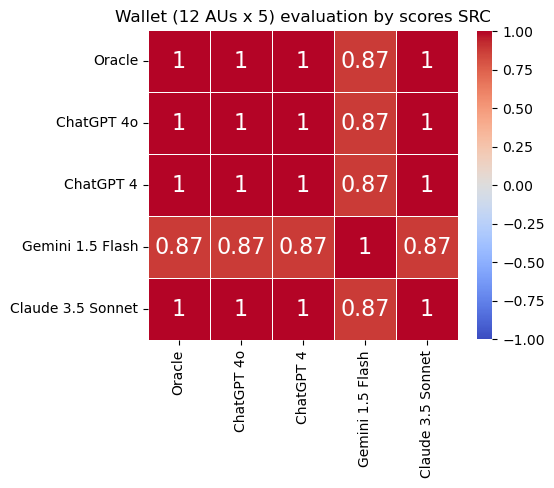} \\ \hline
\textbf{Segmented evaluation by ranking\newline (12 Alternative Uses x 5)} & 
\vspace{2px}\includegraphics[width=0.35\textwidth, height=0.95\textheight, keepaspectratio]{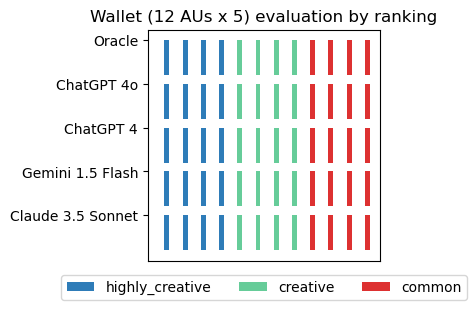} & 
\vspace{2px}\includegraphics[width=0.35\textwidth, height=0.95\textheight, keepaspectratio]{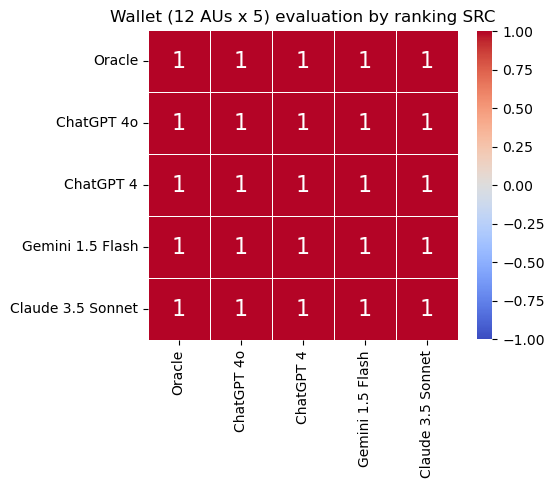} \\ \hline
\end{tabular}
\end{adjustbox}
\vspace{4pt}
\caption{The Alternative Uses orders obtained in the four experiments for a \textbf{wallet}, where every alternative use is colour coded with its creativity category. With the Spearman's Rank Correlation heatmaps.}
\label{tab:wallet_diagrams}
\end{table}

\begin{table}[h!]
\centering
\begin{adjustbox}{max width=\textwidth}
\begin{tabular}{|m{2.2cm}|m{6cm}|m{6cm}|}
\hline
\textbf{Experiments} & \textbf{AUs evaluation order} & \textbf{Spearman's Rank Correlation Heatmap} \\ \hline
\textbf{Comprehensive evaluation by scores\newline (60 Alternative Uses)} & 
\vspace{2px}\includegraphics[width=0.35\textwidth, height=0.95\textheight, keepaspectratio]{Soap/Soap_60_AUs_by_score.png} & 
\vspace{2px}\includegraphics[width=0.35\textwidth, height=0.95\textheight, keepaspectratio]{Soap/Soap_60_AUs_by_score_Heatmap.png} \\ \hline
\textbf{Comprehensive evaluation by ranking\newline (60 Alternative Uses)} & 
\vspace{2px}\includegraphics[width=0.35\textwidth, height=0.95\textheight, keepaspectratio]{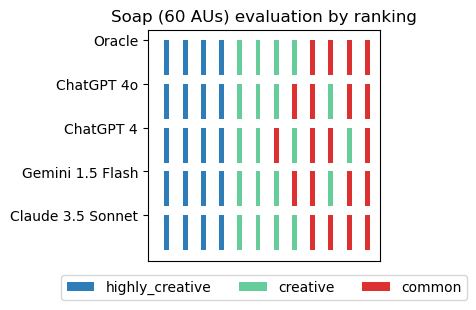} & 
\vspace{2px}\includegraphics[width=0.35\textwidth, height=0.95\textheight, keepaspectratio]{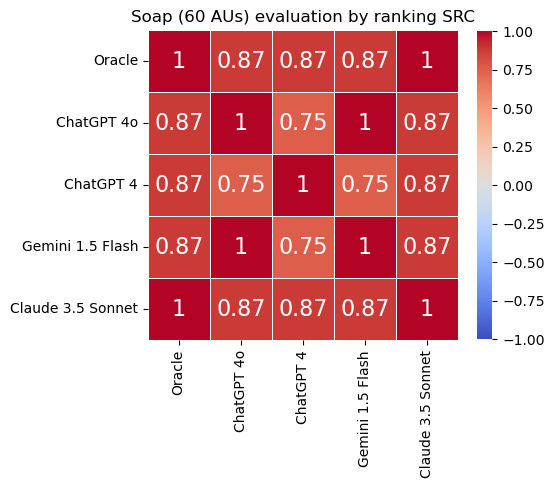} \\ \hline
\textbf{Segmented evaluation by scores\newline (12 Alternative Uses x 5)} & 
\vspace{2px}\includegraphics[width=0.35\textwidth, height=0.95\textheight, keepaspectratio]{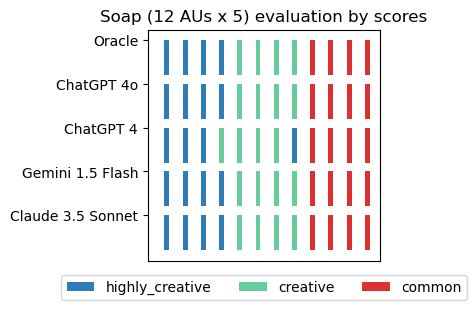} & 
\vspace{2px}\includegraphics[width=0.35\textwidth, height=0.95\textheight, keepaspectratio]{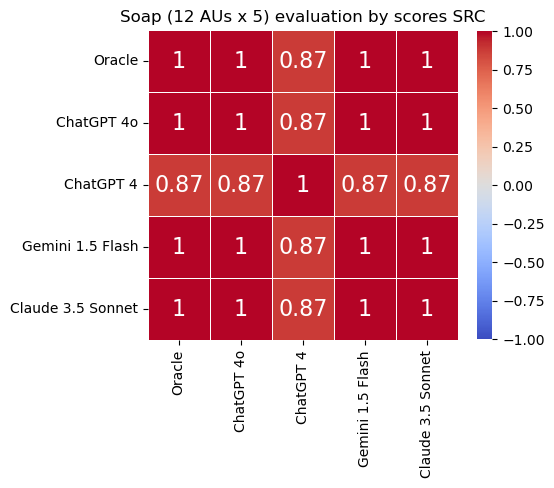} \\ \hline
\textbf{Segmented evaluation by ranking\newline (12 Alternative Uses x 5)} & 
\vspace{2px}\includegraphics[width=0.35\textwidth, height=0.95\textheight, keepaspectratio]{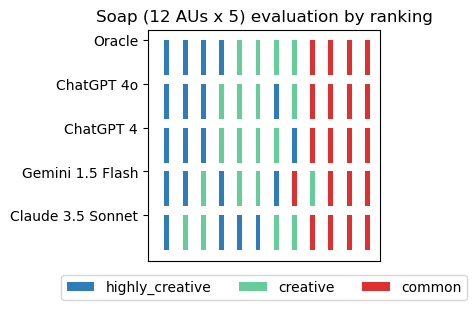} & 
\vspace{2px}\includegraphics[width=0.35\textwidth, height=0.95\textheight, keepaspectratio]{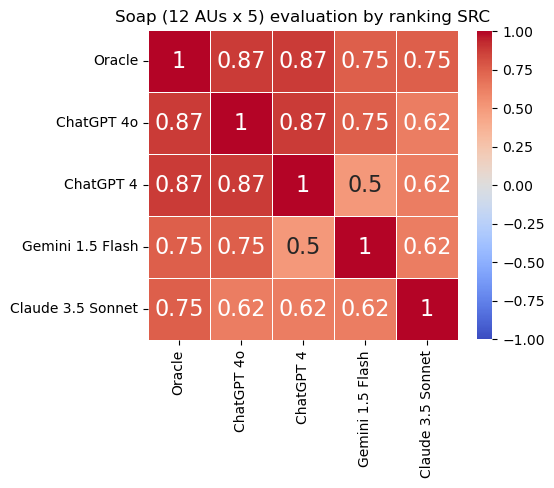} \\ \hline
\end{tabular}
\end{adjustbox}
\vspace{4pt}
\caption{The Alternative Uses orders obtained in the four experiments for a \textbf{soap}, where every alternative use is colour coded with its creativity category. With the Spearman's Rank Correlation heatmaps.}
\label{tab:soap_diagrams}
\end{table}

\begin{table}[h!]
\centering
\begin{adjustbox}{max width=\textwidth}
\begin{tabular}{|m{2.2cm}|m{6cm}|m{6cm}|}
\hline
\textbf{Experiments} & \textbf{AUs evaluation order} & \textbf{Spearman's Rank Correlation Heatmap} \\ \hline
\textbf{Comprehensive evaluation by scores\newline (60 Alternative Uses)} & 
\vspace{2px}\includegraphics[width=0.35\textwidth, height=0.95\textheight, keepaspectratio]{Cotton_swab/Cotton_swab_60_AUs_by_score.png} & 
\vspace{2px}\includegraphics[width=0.35\textwidth, height=0.95\textheight, keepaspectratio]{Cotton_swab/Cotton_swab_60_AUs_by_score_Heatmap.png} \\ \hline
\textbf{Comprehensive evaluation by ranking\newline (60 Alternative Uses)} & 
\vspace{2px}\includegraphics[width=0.35\textwidth, height=0.95\textheight, keepaspectratio]{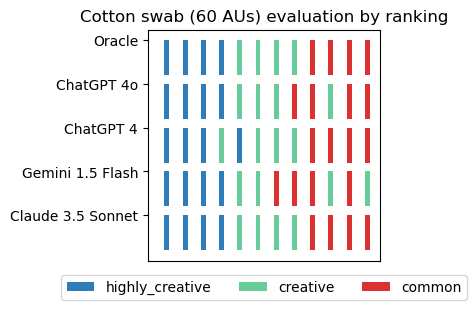} & 
\vspace{2px}\includegraphics[width=0.35\textwidth, height=0.95\textheight, keepaspectratio]{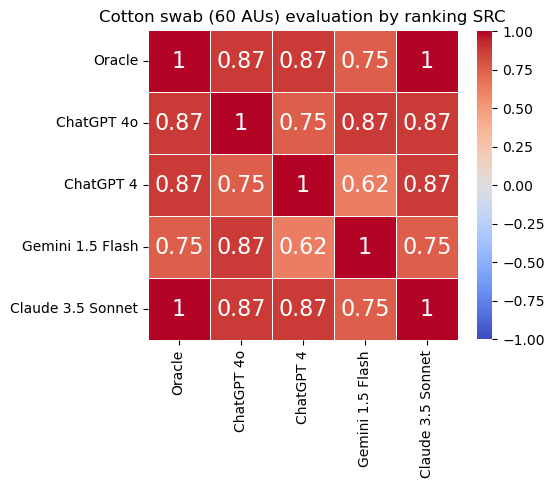} \\ \hline
\textbf{Segmented evaluation by scores\newline (12 Alternative Uses x 5)} & 
\vspace{2px}\includegraphics[width=0.35\textwidth, height=0.95\textheight, keepaspectratio]{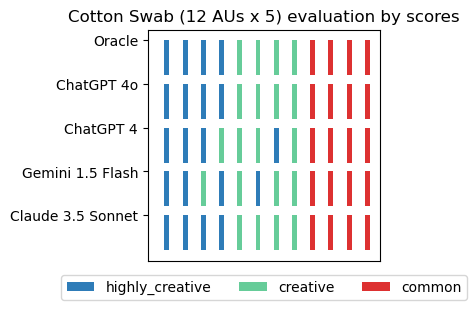} & 
\vspace{2px}\includegraphics[width=0.35\textwidth, height=0.95\textheight, keepaspectratio]{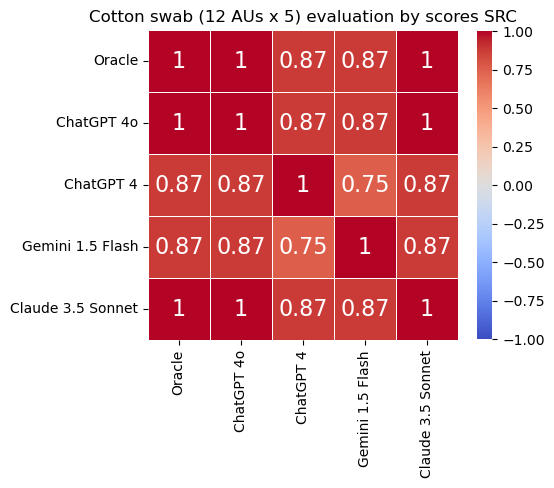} \\ \hline
\textbf{Segmented evaluation by ranking\newline (12 Alternative Uses x 5)} & 
\vspace{2px}\includegraphics[width=0.35\textwidth, height=0.95\textheight, keepaspectratio]{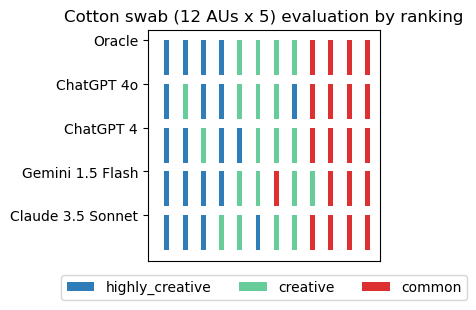} & 
\vspace{2px}\includegraphics[width=0.35\textwidth, height=0.95\textheight, keepaspectratio]{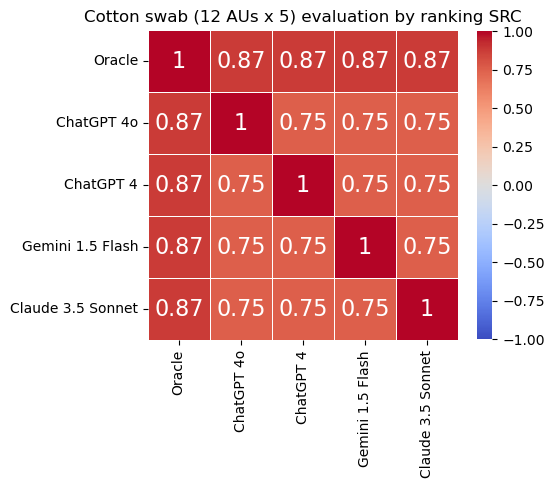} \\ \hline
\end{tabular}
\end{adjustbox}
\vspace{4pt}
\caption{The Alternative Uses orders obtained in the four experiments for a \textbf{cotton swab}, where every alternative use is colour coded with its creativity category. With the Spearman's Rank Correlation heatmaps.}
\label{tab:cotton_swab_diagrams}
\end{table}

\begin{table}[h!]
\centering
\begin{adjustbox}{max width=\textwidth}
\begin{tabular}{|m{2.2cm}|m{6cm}|m{6cm}|}
\hline
\textbf{Experiments} & \textbf{AUs evaluation order} & \textbf{Spearman's Rank Correlation Heatmap} \\ \hline
\textbf{Comprehensive evaluation by scores\newline (60 Alternative Uses)} & 
\vspace{2px}\includegraphics[width=0.35\textwidth, height=0.95\textheight, keepaspectratio]{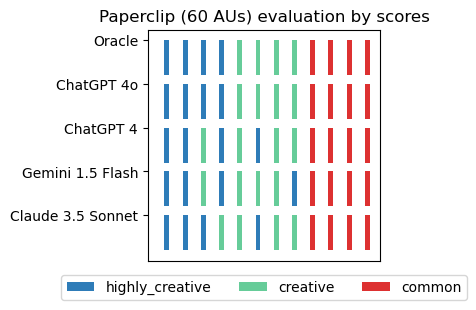} & 
\vspace{2px}\includegraphics[width=0.35\textwidth, height=0.95\textheight, keepaspectratio]{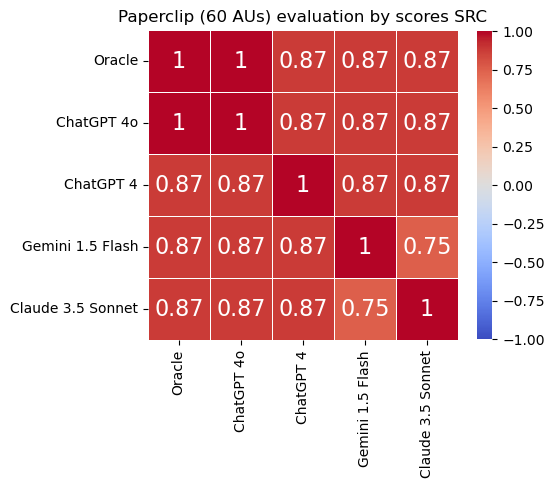} \\ \hline
\textbf{Comprehensive evaluation by ranking\newline (60 Alternative Uses)} & 
\vspace{2px}\includegraphics[width=0.35\textwidth, height=0.95\textheight, keepaspectratio]{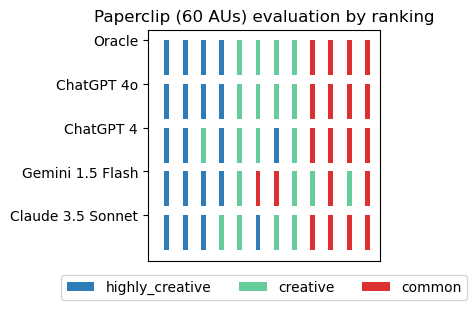} & 
\vspace{2px}\includegraphics[width=0.35\textwidth, height=0.95\textheight, keepaspectratio]{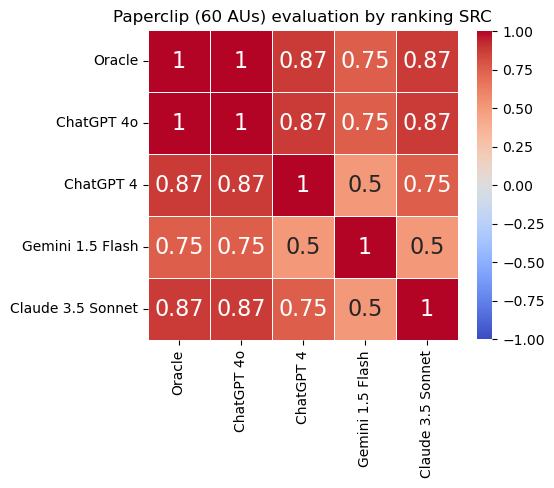} \\ \hline
\textbf{Segmented evaluation by scores\newline (12 Alternative Uses x 5)} & 
\vspace{2px}\includegraphics[width=0.35\textwidth, height=0.95\textheight, keepaspectratio]{Paperclip/Paperclip_12_AUs_by_score.png} & 
\vspace{2px}\includegraphics[width=0.35\textwidth, height=0.95\textheight, keepaspectratio]{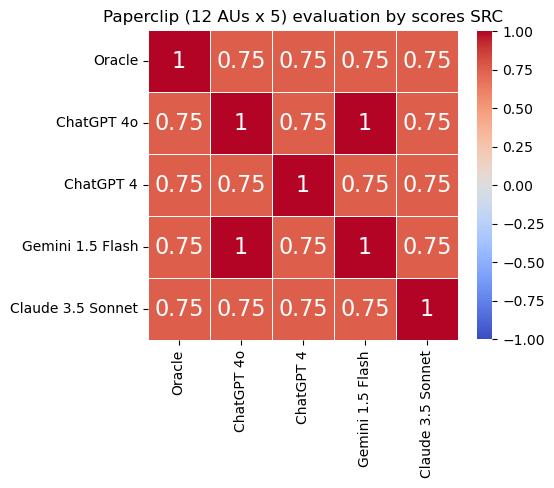} \\ \hline
\textbf{Segmented evaluation by ranking\newline (12 Alternative Uses x 5)} & 
\vspace{2px}\includegraphics[width=0.35\textwidth, height=0.95\textheight, keepaspectratio]{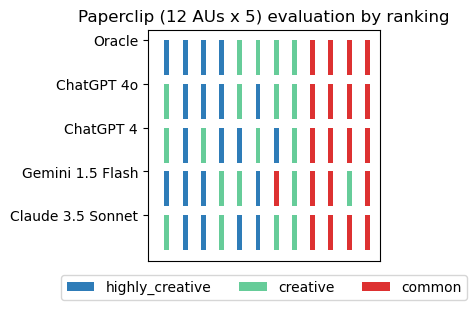} & 
\vspace{2px}\includegraphics[width=0.35\textwidth, height=0.95\textheight, keepaspectratio]{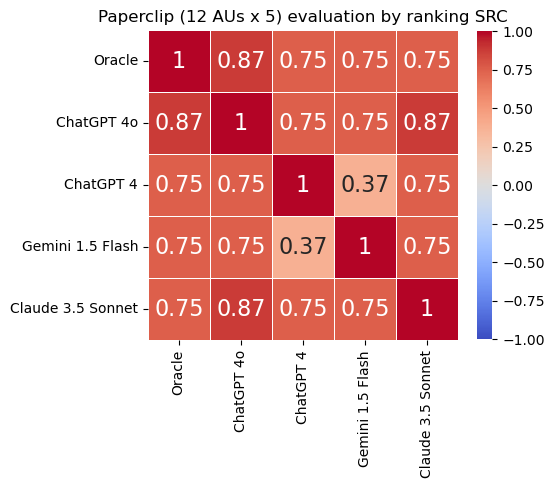} \\ \hline
\end{tabular}
\end{adjustbox}
\vspace{4pt}
\caption{The Alternative Uses orders obtained in the four experiments for a \textbf{paperclip}, where every alternative use is colour coded with its creativity category. With the Spearman's Rank Correlation heatmaps.}
\label{tab:paperclip_diagrams}
\end{table}

\begin{table}[h!]
\centering
\resizebox{\textwidth}{!}{%
\begin{tabular}{@{}l|c|c|c|c|c|c@{}}
\toprule
\textbf{AUs generated by LLMs} & \textbf{\makecell{ChatGPT-4\\evaluation}} & \textbf{\makecell{ChatGPT-4o\\evaluation}} & \textbf{\makecell{Gemini 1.5 Flash\\evaluation}} & \textbf{\makecell{Claude 3.5 Sonnet\\evaluation}} & \textbf{\makecell{Evaluation\\average}} & \textbf{Std. Dev.} \\ \midrule
\multicolumn{7}{c}{All Categories} \\
\midrule
Claude 3.5 Sonnet & 3.19 & 3.17 & 3.24 & 3.00 & \textbf{3.15} & 0.09 \\
ChatGPT-4 & 2.57 & 2.71 & 2.67 & 2.40 & 2.59 & 0.12 \\
Gemini 1.5 Flash & 2.92 & 2.81 & 2.95 & 2.69 & 2.84 & 0.10 \\
ChatGPT-4o & 2.77 & 2.55 & 2.89 & 2.51 & 2.68 & 0.16 \\
\midrule
\multicolumn{7}{c}{common averages} \\
\midrule
Claude 3.5 Sonnet & 1.6 & 1.4 & 1.4 & 1.3 & 1.43 & 0.11 \\
ChatGPT-4 & 1.7 & 1.6 & 1.4 & 1.4 & \textbf{1.53} & 0.13 \\
Gemini 1.5 Flash & 1.7 & 1.4 & 1.4 & 1.4 & 1.48 & 0.13 \\
ChatGPT-4o & 1.7 & 1.4 & 1.5 & 1.4 & 1.50 & 0.12 \\
\midrule
\multicolumn{7}{c}{creative averages} \\
\midrule
Claude 3.5 Sonnet & 3.3 & 3.2 & 3.4 & 2.9 & \textbf{3.20} & 0.19 \\
ChatGPT-4 & 2.9 & 3.1 & 3.1 & 2.6 & 2.93 & 0.20 \\
Gemini 1.5 Flash & 3.0 & 2.8 & 3.0 & 2.6 & 2.85 & 0.17 \\
ChatGPT-4o & 2.7 & 2.5 & 2.9 & 2.3 & 2.60 & 0.22 \\
\midrule
\multicolumn{7}{c}{highly\_creative averages} \\
\midrule
Claude 3.5 Sonnet & 4.6 & 4.9 & 5.0 & 4.8 & \textbf{4.83} & 0.15 \\
ChatGPT-4 & 3.2 & 3.4 & 3.4 & 3.2 & 3.30 & 0.10 \\
Gemini 1.5 Flash & 4.1 & 4.3 & 4.5 & 4.2 & 4.28 & 0.15 \\
ChatGPT-4o & 3.9 & 3.8 & 4.2 & 3.8 & 3.93 & 0.16 \\
\bottomrule
\end{tabular}%
}
\vspace{4pt}
\caption{Average evaluation scores of LLMs' alternative uses for \textbf{evaluation by score (60 AUs)} experiment from 1 (least creative use) to 5 (most creative use).}
\label{tab:evaluation_60AUs_scores}
\end{table}

\begin{table}[h!]
\centering
\resizebox{\textwidth}{!}{%
\begin{tabular}{@{}l|c|c|c|c|c|c@{}}
\toprule
\textbf{AUs generated by LLMs} & \textbf{\makecell{ChatGPT-4\\evaluation}} & \textbf{\makecell{ChatGPT-4o\\evaluation}} & \textbf{\makecell{Gemini 1.5 Flash\\evaluation}} & \textbf{\makecell{Claude 3.5 Sonnet\\evaluation}} & \textbf{\makecell{Evaluation\\average}} & \textbf{Std. Dev.} \\ \midrule
\multicolumn{7}{c}{All Categories} \\
\midrule
Claude 3.5 Sonnet & 5.00 & 5.93 & 5.87 & 5.73 & \textbf{5.63} & 0.37 \\
ChatGPT-4 & 7.13 & 6.13 & 6.53 & 6.87 & 6.67 & 0.38 \\
Gemini 1.5 Flash & 6.87 & 6.93 & 6.47 & 6.67 & 6.74 & 0.18 \\
ChatGPT-4o & 7.00 & 7.00 & 7.13 & 6.73 & 6.97 & 0.15 \\
\midrule
\multicolumn{7}{c}{common averages} \\
\midrule
Claude 3.5 Sonnet & 9.2 & 10.2 & 9.0 & 10.4 & \textbf{9.70} & 0.61 \\
ChatGPT-4 & 10.4 & 8.8 & 9.2 & 11.2 & 9.90 & 0.95 \\
Gemini 1.5 Flash & 11.4 & 11.8 & 10.2 & 10.4 & 10.95 & 0.67 \\
ChatGPT-4o & 10.0 & 10.4 & 9.0 & 10.0 & 9.85 & 0.52 \\
\midrule
\multicolumn{7}{c}{creative averages} \\
\midrule
Claude 3.5 Sonnet & 4.4 & 6.4 & 7.2 & 5.6 & 5.90 & 1.03 \\
ChatGPT-4 & 6.4 & 4.6 & 5.6 & 5.4 & \textbf{5.50} & 0.64 \\
Gemini 1.5 Flash & 7.2 & 6.8 & 7.2 & 7.4 & 7.15 & 0.22 \\
ChatGPT-4o & 7.6 & 6.8 & 8.6 & 7.0 & 7.50 & 0.70 \\
\midrule
\multicolumn{7}{c}{highly\_creative averages} \\
\midrule
Claude 3.5 Sonnet & 1.4 & 1.2 & 1.4 & 1.2 & \textbf{1.30} & 0.10 \\
ChatGPT-4 & 4.6 & 5.0 & 4.8 & 4.0 & 4.60 & 0.37 \\
Gemini 1.5 Flash & 2.0 & 2.2 & 2.0 & 2.2 & 2.10 & 0.10 \\
ChatGPT-4o & 3.4 & 3.8 & 3.8 & 3.2 & 3.55 & 0.26 \\
\bottomrule
\end{tabular}%
}
\vspace{4pt}
\caption{Average evaluation ranking of LLMs' alternative uses for \textbf{evaluation by ranking (60 AUs)} experiment from 1 (most creative use) to 12 (least creative use) as we took the ranking average for each 5 alternative uses with the same model and category.}
\label{tab:evaluation_60AUs_ranking}
\end{table}

\begin{table}[h!]
\centering
\resizebox{\textwidth}{!}{%
\begin{tabular}{@{}l|c|c|c|c|c|c@{}}
\toprule
\textbf{AUs generated by LLMs} & \textbf{\makecell{ChatGPT-4\\evaluation}} & \textbf{\makecell{ChatGPT-4o\\evaluation}} & \textbf{\makecell{Gemini 1.5 Flash\\evaluation}} & \textbf{\makecell{Claude 3.5 Sonnet\\evaluation}} & \textbf{\makecell{Evaluation\\average}} & \textbf{Std. Dev.} \\ \midrule
\multicolumn{7}{c}{All Categories} \\
\midrule
Claude 3.5 Sonnet & 3.0 & 3.2 & 2.8 & 3.0 & \textbf{3.00} & 0.14 \\
ChatGPT-4 & 2.8 & 3.1 & 2.7 & 2.8 & 2.85 & 0.15 \\
Gemini 1.5 Flash & 2.8 & 2.9 & 2.4 & 2.7 & 2.70 & 0.19 \\
ChatGPT-4o & 3.0 & 3.0 & 2.6 & 2.8 & 2.85 & 0.17 \\
\midrule
\multicolumn{7}{c}{common averages} \\
\midrule
Claude 3.5 Sonnet & 1.4 & 1.6 & 1.6 & 1.4 & 1.50 & 0.10 \\
ChatGPT-4 & 1.6 & 1.8 & 1.7 & 1.4 & \textbf{1.63} & 0.15 \\
Gemini 1.5 Flash & 1.5 & 1.5 & 1.5 & 1.4 & 1.48 & 0.04 \\
ChatGPT-4o & 1.6 & 1.6 & 1.4 & 1.5 & 1.53 & 0.08 \\
\midrule
\multicolumn{7}{c}{creative averages} \\
\midrule
Claude 3.5 Sonnet & 3.6 & 3.7 & 3.2 & 3.6 & \textbf{3.53} & 0.19 \\
ChatGPT-4 & 3.3 & 3.5 & 3.1 & 3.4 & 3.33 & 0.15 \\
Gemini 1.5 Flash & 3.3 & 3.5 & 2.6 & 3.1 & 3.13 & 0.33 \\
ChatGPT-4o & 3.0 & 3.0 & 2.4 & 2.8 & 2.80 & 0.24 \\
\midrule
\multicolumn{7}{c}{highly\_creative averages} \\
\midrule
Claude 3.5 Sonnet & 4.1 & 4.3 & 3.6 & 4.1 & 4.03 & 0.26 \\
ChatGPT-4 & 3.4 & 4.1 & 3.2 & 3.6 & 3.58 & 0.33 \\
Gemini 1.5 Flash & 3.7 & 3.8 & 3.0 & 3.6 & 3.53 & 0.31 \\
ChatGPT-4o & 4.4 & 4.4 & 4.0 & 4.2 & \textbf{4.25} & 0.17 \\
\bottomrule
\end{tabular}%
}
\vspace{4pt}
\caption{Average evaluation scores of LLMs' alternative uses for \textbf{evaluation by score (12 AUs x 5)} experiment from 1 (least creative use) to 5 (most creative use).}
\label{tab:evaluation_12AUs_scores}
\end{table}

\begin{table}[h!]
\centering
\resizebox{\textwidth}{!}{%
\begin{tabular}{@{}l|c|c|c|c|c|c@{}}
\toprule
\textbf{AUs generated by LLMs} & \textbf{\makecell{ChatGPT-4\\evaluation}} & \textbf{\makecell{ChatGPT-4o\\evaluation}} & \textbf{\makecell{Gemini 1.5 Flash\\evaluation}} & \textbf{\makecell{Claude 3.5 Sonnet\\evaluation}} & \textbf{\makecell{Evaluation\\average}} & \textbf{Std. Dev.} \\ \midrule
\multicolumn{7}{c}{All Categories} \\
\midrule
Claude 3.5 Sonnet & 6.1 & 6.2 & 5.9 & 6.3 & \textbf{6.13} & 0.15 \\
ChatGPT-4 & 6.6 & 6.4 & 6.7 & 6.5 & 6.55 & 0.11 \\
Gemini 1.5 Flash & 6.8 & 7.0 & 6.8 & 6.8 & 6.85 & 0.09 \\
ChatGPT-4o & 6.5 & 6.5 & 6.6 & 6.3 & 6.48 & 0.11 \\
\midrule
\multicolumn{7}{c}{common averages} \\
\midrule
Claude 3.5 Sonnet & 9.6 & 9.6 & 8.5 & 10.2 & 9.48 & 0.61 \\
ChatGPT-4 & 9.7 & 8.8 & 9.2 & 9.5 & \textbf{9.30} & 0.34 \\
Gemini 1.5 Flash & 10.2 & 10.2 & 10.1 & 10.0 & 10.13 & 0.08 \\
ChatGPT-4o & 10.1 & 10.1 & 9.6 & 9.8 & 9.90 & 0.21 \\
\midrule
\multicolumn{7}{c}{creative averages} \\
\midrule
Claude 3.5 Sonnet & 4.6 & 4.8 & 5.4 & 5.0 & \textbf{4.95} & 0.30 \\
ChatGPT-4 & 5.3 & 5.3 & 6.0 & 5.0 & 5.40 & 0.37 \\
Gemini 1.5 Flash & 5.6 & 5.8 & 5.9 & 5.8 & 5.78 & 0.11 \\
ChatGPT-4o & 6.9 & 6.2 & 7.0 & 6.0 & 6.53 & 0.43 \\
\midrule
\multicolumn{7}{c}{highly\_creative averages} \\
\midrule
Claude 3.5 Sonnet & 3.9 & 4.0 & 3.9 & 3.6 & 3.85 & 0.15 \\
ChatGPT-4 & 4.9 & 5.0 & 4.8 & 5.0 & 4.93 & 0.08 \\
Gemini 1.5 Flash & 4.7 & 5.0 & 4.4 & 4.6 & 4.68 & 0.22 \\
ChatGPT-4o & 2.4 & 3.1 & 3.2 & 3.2 & \textbf{2.98} & 0.33 \\
\bottomrule
\end{tabular}%
}
\vspace{4pt}
\caption{Average evaluation ranking of LLMs' alternative uses for \textbf{evaluation by ranking (12 AUs x 5)} experiment from 1 (most creative use) to 12 (least creative use) as we took the ranking average for each 5 alternative uses with the same model and category.}
\label{tab:evaluation_12AUs_ranking}
\end{table}
\end{document}